\documentclass[tablecaption=bottom,pmlr]{jmlr}

\usepackage[utf8]{inputenc} 
\usepackage[T1]{fontenc}    
\usepackage{hyperref}       
\usepackage{url}            
\usepackage{booktabs}       
\usepackage{amsfonts}       
\usepackage{amsmath}
\usepackage{nicefrac}       
\usepackage{microtype}      

\jmlrvolume{148}
\jmlryear{2021}
\jmlrsubmitted{submission date}
\jmlrpublished{publication date}
\jmlrworkshop{NeurIPS 2020 Preregistration Workshop} 

\title{Testing the Genomic Bottleneck Hypothesis in Hebbian Meta-Learning}

\author{\Name{Rasmus Berg Palm} \Email{rasmb@itu.dk}\\
\Name{Elias Najarro} \Email{enaj@itu.dk}\\
\Name{Sebastian Risi} \Email{sebr@itu.dk}\\
\addr IT University of Copenhagen}

\begin{document}

\maketitle

\begin{abstract}
Hebbian meta-learning has recently shown promise to solve hard reinforcement learning problems, allowing agents to adapt to some degree to changes in the environment. However, because each synapse in these approaches can learn a very specific learning rule, the ability to generalize to very different situations is likely reduced. 
  We hypothesize that limiting the number of Hebbian learning rules through a ``genomic bottleneck'' can act as a regularizer leading to better generalization across changes to the environment. We test this hypothesis by decoupling the number of Hebbian learning rules from the number of synapses and systematically varying the number of Hebbian learning rules. 
  The results in this paper suggest that simultaneously learning the Hebbian learning rules and their assignment to synapses is a difficult optimization problem, leading to poor performance in the environments tested. However, parallel research to ours finds that it is indeed possible to reduce the number of learning rules by clustering similar rules together. How to best implement a ``genomic bottleneck'' algorithm is thus an important research direction that warrants further investigation.
\end{abstract}


\section{Introduction}

Deep reinforcement learning has made great progress recently on very hard problems like Go, Starcraft and Dota using deep neural networks \citep{silver2016mastering, vinyals2019grandmaster, berner2019dota}. However, once learned, the networks are static and highly specific. As such, there is very little capacity to adapt to changes in the environment or to generalize across environments \citep{justesen2018illuminating}. For instance, the state-of-the-art AlphaStar agent trained to play one race in Starcraft cannot play another, even though it is a very similar task, and much less play Go or load a dishwasher. In contrast animals and humans show remarkable flexibility in their ability to generalize across tasks and adapt to changes.

Meta-learning proposes to overcome these limitations by learning-to-learn. That is to learn general learning rules that are broadly applicable and enable an agent to quickly adapt to changes in the environment or new tasks \citep{finn2017model, wang2016learning, weng2018metalearning}.

One particularly interesting approach to meta-learning is Hebbian meta-learning. The goal in Hebbian meta-learning is to learn Hebbian learning rules that enable an agent to quickly learn to perform well in its environment and adapt to changes. Hebbian learning is a learning paradigm in which the synaptic strength between neurons is determined by the correlation of their activity \citep{hebb2005organization}; informally, 
``neurons that fire together, wire together''. The Hebbian learning paradigm is promising since it proposes a simple and \emph{general} learning paradigm supported by extensive empirical evidence in biological brains \citep{prezioso2018spike, caporale2008spike}.

\citet{najarro2020meta} showed promising results using Hebbian meta-learning to solve a difficult car racing reinforcement problem and quadruped locomotion. Impressively the agent's policy network was initialized randomly each episode, so it had to learn the task during its lifetime using only the meta-learned Hebbian learning rules. Further, the learned Hebbian learning rules generalized to an unseen quadruped leg damage scenario, which baseline non-plastic feedforward neural networks could not. However, each synapse had its own learning rule, which allowed the network to learn very specific learning rules for each neuron. This raises the question to which degree the network learned to learn, or whether each learning rule rather encoded very specific dynamics for each synapse.

We hypothesize that the architecture in \citet{najarro2020meta} is limited in its ability to discover general learning rules, but rather encodes very specific dynamics for each synapse. We further hypothesize that in order to learn generally useful learning rules the network must be limited to how many learning rules it can learn. This is inspired by the genomic bottleneck observed in humans and complex animals, where the information that can be stored in the genome is several orders of magnitude smaller than what is needed to determine the final wiring of the brain \citep{zador2019critique}. We hypothesize that limiting the number of learning rules acts as a regularizer, which improves generalization and adaptability.

\section{Related Work}
 
\textbf{Meta-Learning}. In meta-learning, the goal is to learn to quickly adapt to a target task given a set of training tasks \citep{weng2018metalearning}. Common approaches can loosely be categorized into black-box, optimization, and metric-based. Black-box methods learn a function that, conditioned on samples from a new task, outputs a function to solve the new task. For instance by jointly learning a network that can produce a latent summary of a new task and a network that can solve the task given the latent summary \citep{santoro2016meta, mishra2017simple}. Optimization-based attempts to learn to quickly optimize on a new task, e.g.\ by finding an initialization from which optimization on new tasks is fast \citep{finn2017model, nichol2018first} or by learning the optimization algorithm \citep{andrychowicz2016learning, li2017meta}. Metric based approaches learns an embedding space that facilitates effective distance-based classification, e.g.\ Siamese networks \citep{koch2015siamese} or prototypical networks \citep{snell2017prototypical}. Meta Reinforcement Learning (meta-RL) extends the meta-learning idea to the reinforcement learning setting. Formally the goal is to quickly learn to perform well in a new Markov Decision Process (MDP) given a set of training MDPs. A common approach is to learn a recurrent neural network policy where the hidden activations are not reset between episodes \citep{wang2016learning, duan2016rl}. This allows the policy network to discover how the environment behaves across episodes and adapt its policy. Another common approach is meta-learning an initialization of a policy network from which policy gradient descent can quickly adapt to the new MDP \citep{finn2017model, song2019maml}.

\textbf{Plastic Artificial Neural Networks} A less explored meta-learning approach is based on plastic neural networks that are optimized to have both innate properties and the ability to learn during their lifetime. For example, such networks can learn by changing the connectivity among neurons through local learning rules like Hebbian plasticity  \citep{soltoggio2018born,soltoggio2007evolving}.  Often these networks are optimized through evolutionary algorithms \citep{soltoggio2018born, najarro2020meta} but more recently optimizing the plasticity of connections in a network through gradient descent has also been shown possible \citep{miconi2018differentiable}. However, in contrast to the evolving Hebbian learning rules approach in \citet{najarro2020meta}, the gradient descent approach \citep{miconi2018differentiable} was so far restricted to only evolving a single plasticity parameter for each connection instead of a different Hebbian rule.

\textbf{Fast Weights} Artificial neural networks (ANN) have either a slow form of storing information---through updating the weights---or, if they are recurrent, a very fast form of information storage in the form of internal activations. Fast weights seek to introduce an intermediate time-scale to the information storage in ANNs \citep{hinton86, huber92} and is motivated by the observation that biological neural networks have learning processes occurring concurrently that span across very different time scales. Recently, this approach has been successfully applied to image recognition tasks as well as a model for content-addressable memory \citep{Ba2016Oct}. In the context of meta-learning, fast weights has been shown to perform meta-RL by having an ANN update the weights of a policy network on a per-task basis \citep{Munkhdalai2017}. Additionally, \citet{munkhdalai2018} showed that a fast-weights Hebbian mechanism is capable of performing one-shot supervised learning tasks.

\section{Hebbian meta-learning}

In Hebbian meta-learning, the goal is to meta-learn Hebbian learning rules that enable an agent to perform well, and adapt to changes, in its environment.

\subsection{Hebbian Learning}

The agent acts in an episodic reinforcement learning environment and is controlled by a neural policy network. At the start of each episode, the policy network is initialized with random weights, which then undergoes changes according to Hebbian learning rules during the episode. Specifically, we use the ABCD Hebbian weight plasticity formalization \citep{soltoggio2007evolving} and the weights are updated at each step of the episode. The change to the weight connecting neuron $i$ and $j$ is
\begin{align}
    \label{eq:hebbian}
    \Delta w_{ij} = \eta_{ij} \left(A_{ij} o_i o_j + B_{ij} o_i + C_{ij} o_j + D_{ij}\right) \,,
\end{align}
where $o_i$ and $o_j$ are the pre- and post-synaptic activation of the neurons and  $h = \{\eta, A, B, C, D\}$ are the Hebbian parameters that are fixed during the episode. 

\subsection{Evolutionary Strategies}
The Hebbian parameters are meta-learned on an evolutionary time scale $t$, using evolutionary strategies (ES) \citep{salimans2017evolution}. In evolutionary strategies the goal is to maximize the expected fitness of a distribution of individuals, $\max_\theta \mathbb{E}_{z \sim p(z|\theta)} F(z)$, where $F(z)$ is the fitness of an individual $z$ and $\theta$ parameterize this distribution. We compute the gradient of this objective using the score function estimator and use gradient ascent to maximize it,
\begin{align*}
\nabla_\theta \mathbb{E}_{z\sim p(z|\theta)} F(z) &= \mathbb{E}_{z\sim p(z|\theta)} F(z) \nabla_\theta \log(p(z|\theta))
\end{align*}
\begin{align}
\label{eq:es}
\theta^{t+1} &= \theta^t + \alpha \left[\mathbb{E}_{z\sim p(z|\theta^t)} F(z) \nabla_{\theta^t} \log(p(z|\theta^t))\right] \,,
\end{align}
where $\alpha$ is the learning rate. The expectation is evaluated using $n$ samples, the population size. In order to use ES then, we must define $F(z)$ and $p(z|\theta)$. In this paper $F(z)$ is always the accumulated reward of an episode.

\subsection{Individual Learning Rules}
For the case of individual learning rules, $N$ synapses each have their own learning rules $h_i = [\eta_i, A_i, B_i, C_i, D_i]$,  which are drawn from independent normal distributions with fixed variance $\sigma^2$ and meta-learned means $\mu \in \mathbb{R}^{N \times 5}$. In this case $p(z|\theta) = p(h|\mu) = \prod_{i=1}^N \mathcal{N}(h_i|\mu_i, \sigma)$, where $\mu_i$ denotes the $i$'th row of the $\mu$ parameters. Inserting into eq. (\ref{eq:es}) and deriving the gradient this reduces to the expression in \citep{najarro2020meta, salimans2017evolution},
\begin{align*}
    \mu^{t+1} = \mu^t + \alpha\left[ \frac{1}{\sigma} \mathbb{E}_{\epsilon \sim \mathcal{N}(0, I)} F(\mu^t + \sigma \epsilon) \epsilon \right] \,,
\end{align*}
where $\epsilon$ is a sample from a standard normal $\mathcal{N}(0,I)$.

\subsection{Shared Learning Rules}

To explore the effect of sharing a limited amount of Hebbian learning rules we sample each synapses' learning rule from a Gaussian Mixture Model (GMM) with $M$ clusters, such that
\begin{align}
  k &\sim p(k|\lambda) \in [1,...,M]^N \,, \lambda \in \mathbb{R}^{N \times M} \,, \nonumber\\
  h &\sim \mathcal{N}(h|\mu_k, \sigma) \in \mathbb{R}^{N \times 5} \,, \mu \in \mathbb{R}^{M \times 5} \,, \nonumber\\
  p(z|\theta) = p(h|\mu, \lambda) &= \prod_{i=1}^N \sum_{k=1}^M \mathcal{N}(h_i | \mu_{k}, \sigma) p(k|\lambda_i) \,.
  \label{eq:gmm_prob}
\end{align}
Here $ p(k|\lambda_i) = e^{\lambda_{ik}} / \sum_{j=1}^M e^{\lambda_{ij}}$, i.e. the softmax categorical distribution parameterized by $\lambda_i$ logits for synapse $i$ and the meta learned parameters are $\mu$ and $\lambda$. We use automatic differentiation to compute the gradient of the log likelihood of this distribution with respect to $\mu$ and $\lambda$, and then use eq. (\ref{eq:es}) to update $\mu$ and $\lambda$.

This approach to assigning learning rules to synapses is the most flexible and direct but requires $N \times M$ parameters. However, we are only interested in testing the effect of limiting the number of learning rules, not the number of parameters or bits needed to encode an individual, although those are interesting directions for future work.

\section{Experiments}

The first experiment is to replicate the original results of \citet{najarro2020meta} to ensure a fair comparison. In all experiments, we use the same experimental setup, architectures, and hyper-parameters as in \citet{najarro2020meta} except where noted.

Similar to \citet{najarro2020meta} we experiment on the car racing and quadruped locomotion tasks. We perform leave-one-out cross-validation, with five variations for each task. The five variations for the car racing tasks are (1) default settings, (2,3) twice and half the road friction coefficient, and (4,5) constant force pushing the car west and east. For the quadruped locomotion task, the variations are (1) default settings, (2,3) left and right front leg damage as in \citet{najarro2020meta} and (4,5) 50\% longer rear and front legs. We perform leave-one-out cross-validation by leaving one variation out for testing and training on the remaining variations. Specifically, we use eq. (\ref{eq:es}) to maximize $\mathbb{E}_{z \sim p(z|\theta)} F'(z)$ where $F'$ is the fitness function of a meta-task formed by taking the average fitness across the four training variations of the task.

To test our core hypothesis we evaluate for a varying number of learning rules expressed as a fraction of the number of synapses such that $M=\frac{1}{\rho} N$. We vary $\rho = [1, 16, 32, 64, 128, 256, N]$, where $\rho=1$ corresponds to a learning rule per synapse and $\rho=N$ corresponds to a single learning rule. 

We compare to three baselines in all experiments: (1) A Hebbian meta-learning network with a learning rule per synapse as in \citet{najarro2020meta}, (2) an identical static network with learned weights, and (3) a LSTM baseline with hidden states initialized to zero at the start of each episode \citep{hochreiter1997long}. We construct the LSTM baseline by replacing the first densely connected layer in the static baseline with a LSTM layer and keeping everything else the same. All architectures are optimized using ES.

\section{Results}

We use the EvoStrat computational library \citep{palm2020} for all experiments. Code to reproduce all results are available at \hyperlink{https://github.com/rasmusbergpalm/hebbian-evolution}{github.com/rasmusbergpalm/hebbian-evolution}.

\subsection{Replication}
We successfully replicated the results of \citet{najarro2020meta} for the car racing and quadruped walker environments consistently achieving high rewards in 300 and 500 generations respectively with a Hebbian network with a learning rule per synapse.

\subsection{Computational issues}
Evaluating eq. \ref{eq:es} for the shared Hebbian learning rules for small values of $\rho$, becomes very expensive in terms of memory due to the number of parameters of the model. Unfortunately, that means we're not able to perform the experiments for values of $\rho<32$.

\subsection{Ant Environment}

The results on the quadruped walker environment "AntBulletEnv-v0" are summarized in table \ref{table:ant-results}. The networks are trained on the 4 left-most morphologies and tested on the "Long front" morphology corresponding to 50\% longer front legs.\\ 
\textbf{Baselines} The baselines all do reasonably well. The static network is best on average on the training tasks, whereas the LSTM and Hebbian learning rules are better on the unseen test task, although the standard deviations are large. This is in line with the original results from \citet{najarro2020meta}.\\
\textbf{Shared learning rules} We first test with $\rho=128$, which fails on both the training and testing morphology. Given that it fails on the training tasks we only evaluate subsequent values of $\rho$ on the default training morphology. We assume that if it fails here it will also fail under the more difficult leave-one-out meta-learning testing scheme.

\begin{table}[!ht]
\caption{Ant environment results. Average distance traveled. Bold indicate unseen testing morphology. Mean and standard deviation from 100 samples. Rounded to nearest integer.}
\label{table:ant-results}
\centering
  \begin{tabular}{lccccc}
  \toprule
    Model   & Default       & Damage left   & Damage right  & Long back     & \textbf{Long front}  \\  
  \midrule
    Static  & $1545 \pm 30$ & $839 \pm 60$   & $1891 \pm 105$ & $1197 \pm 31 $ & $202 \pm 174$\\ 
    LSTM & $1097 \pm 57$ & $1374 \pm 74$ &  $1240 \pm 140$ & $806 \pm 113$ & $314 \pm 115$\\ 
    Hebbian & $928 \pm 88$  & $1102 \pm 164$ & $981 \pm 140$ & $741 \pm 169$ & $364 \pm 131$\\ 
    Shared($\rho=32$) & $2 \pm 12$ &  &  &  & \\ 
    Shared($\rho=64$) & $3 \pm 13$ &  &  &  & \\ 
    Shared($\rho=128$) & $0 \pm 10$ & $-9 \pm 18$ & $15 \pm 31$ & $1 \pm 6$ & $0 \pm 4$\\ 
    Shared($\rho=256$) & $2 \pm 12$ &  &  &  & \\ 
    Shared($\rho=N$) & $-1 \pm 0$ &  &  &  & \\ 
  \bottomrule
  \end{tabular}  
\end{table}

\subsection{Car Racing Environment}

\textbf{Baselines} All the baselines learn to perform well, surprisingly also on the unseen test environment variation.\\
\textbf{Shared learning rules} Similarly to the ant results, we test with $\rho=128$ first, and it fails to learn anything on both the training and testing tasks. Given similar negative results from the ant environment, we conclude that there is something fundamentally problematic about the shared learning rule experiments, and focus on diagnosing that. As such we do not test for other values of $\rho$ on the car environment.

\begin{table}[!ht]
\caption{Car racing environment results. Average distance traveled.}
\label{table:results}
\centering
  \begin{tabular}{lccccc}
  \toprule
    Model   & Default       & Half friction   & Double friction  & Pushed left     & \textbf{Pushed right}  \\  
  \midrule
    Static  & $784 \pm 119$ & $491 \pm 202$   & $789 \pm 202$ & $738 \pm 219 $ & $628 \pm 228$\\ 
    LSTM & $680 \pm 247$ & $375 \pm 209$ &  $711 \pm 243$ & $653 \pm 273$ & $610 \pm 316$\\ 
    Hebbian & $674 \pm 135$  & $552 \pm 190$ & $692 \pm 120$ & $725 \pm 139$ & $606 \pm 186$\\ 
    Shared($\rho=128$) & $6 \pm 3$ & $5 \pm 1$ & $8 \pm 7$ & $6 \pm 1$ & $6 \pm 2$\\ 
  \bottomrule
  \end{tabular}  
\end{table}

\subsection{Post-hoc analysis}

We deviate from our pre-planned experimental setup to see if we can get the shared Hebbian learning rules to work. We experiment on the default Ant morphology using $\rho=128$. We independently try lowering the learning rate to $0.01$ (from $0.2$), reducing the number of shared learning rules to $16$ and initializing the shared learning rule means from $\mathcal{N}(0, 10)$ (from $\mathcal{N}(0, 1)$). However, none of these modifications helped. We also try randomly assigning shared learning rules instead of learning the assignments. This modification improves performance, however, the networks are still much worse than the baselines (not shown). We don't pursue the random assignment any further since we are interested in learning the assignments. 

We note that the Hebbian networks with shared learning rules fail to learn the mixing weights $\lambda$, which stay close to a uniform distribution, and that the gradient estimates on the mixing weights are two orders of magnitude smaller than on the learning rule means $\mu$. We manually verify that the $\log p(h|\mu, \lambda, \sigma)$ computations and the gradients $\nabla_\mu \log p(h|\mu, \lambda, \sigma)$ and $\nabla_\lambda \log p(h|\mu, \lambda, \sigma)$ are correct for a simple test case.

Suspecting a difficult optimization landscape we use the Adam optimizer which scales the learning rate individually for each parameter \citep{kingma2014adam}. With the Adam optimizer, the networks attain poor but non-zero performance. This rules out the most severe bugs in the implementation and experimental setup and lends support to our hypothesis that it is ``just'' a difficult optimization problem. We experiment with several learning rates and learning rate schedules, achieving the best results using a learning rate of $1.0$ and an exponential learning rate decay of $0.9931$ which corresponds to halving the learning rate every 100 updates. However, these results are still much worse than the baselines, achieving $555 \pm 89$ reward on the default Ant morphology.

\section{Discussion}

The experimental results in this paper do not support our initial hypothesis. However, parallel research to ours found that it is indeed possible to reduce the number of Hebbian learning rules significantly and improve generalization on unseen environments by iteratively clustering similar learning rules and re-training \citep{pedersen2021evolving}. As such, the results suggest that our approach (i.e.\ modeling the learning rules as drawn from a Gaussian Mixture Model and jointly optimizing the assignment and the learning rules) is a hard optimization problem. 

There is a bit of a chicken and egg problem in jointly optimizing the assignment and the learning rules: it is challenging to optimize the learning rules before they are assigned to synapses, and it is challenging to optimize the assignment to synapses before the learning rules are learned. The approach introduced by \citet{pedersen2021evolving} avoids this issue by learning a Hebbian learning rule per synapse, while periodically clustering similar learning rules and re-training with the reduced set of learning rules. It could also be possible to avoid this chicken and egg problem by alternately learning the assignment and the learning rules in an Expectation-Maximization (EM) inspired approach \citep{dempster1977maximum}.

Another possible explanation for the difficulty in directly learning a small number of shared Hebbian learning rules is the lottery ticket hypothesis \citep{frankle2018lottery}. The lottery ticket hypothesis is based on the observation that neural networks can often be pruned aggressively without losing performance, and that it is in fact a small sub-network of the bigger network which solves the task; however, it is difficult to learn the small sub-networks directly.
The hypothesis is that overparameterized neural networks are more likely to contain sub-network, which are initialized in such a way that they can be effectively optimized to solve the task. If finding a well-initialized sub-network corresponds to winning the lottery then bigger networks simply have more tickets and are thus more likely to win.

The main positive finding of this paper is that we reproduce the results from \citep{najarro2020meta} with different tasks and a different implementation. We thus add to the growing body of evidence that shows it is indeed possible to solve complex tasks with randomly initialized neural networks that learn using local Hebbian learning rules.

One interesting idea for future work is to re-define an individual $z=\{h, k\}$, i.e. the sampled learning rules $h$ \emph{and} the learning rule clusters they are sampled from $k$, such that $p(z|\theta) = p(h, k|\theta) = \prod_{i=1}^N \mathcal{N}(h_i|\mu_k, \sigma)p(k_i|\lambda_i)$. 
Without the inner sum, this setup would be much cheaper to optimize for many learning rule clusters. 
It might also result in an easier optimization landscape. However, it would technically not be correct since an individual's fitness only depends on the sampled learning rules $h$, not which learning rule cluster it is sampled from $k$.

A limitation of our approach to Hebbian meta-learning is that the agent cannot adapt to changes in the reward function during its lifetime, since it does not observe the reward. The reward is only observed at the evolutionary time scale. As such all the training and test tasks must share the same reward function. This is in contrast to common benchmarks in meta RL where the reward function differs across training and test MDPs. Extending our method with approaches that can modulate Hebbian learning based on the reward \citep{abbott1990modulation, krichmar2008neuromodulatory, soltoggio2007evolving, soltoggio2018born} or indirectly encode plasticity   \citep{risi2010indirectly}, could be a promising
future research direction. 

\textbf{Changes to the pre-registration proposal.} The pre-registration proposal sections (1, 2, 3, and 4) have been changed insignificantly to correct grammar and spelling. The main findings have been added to the abstract.

\bibliography{refs}

\end{document}